\documentclass{article}
\usepackage{amssymb}
\usepackage{changes}
\usepackage{spconf,amsmath,graphicx}
\usepackage{adjustbox}
\usepackage{cleveref}
\usepackage{changes}
\usepackage{multirow}
\usepackage{xcolor}
\usepackage{soul}
\usepackage{colortbl}
\usepackage[tight,small]{subfigure}

\definecolor{addcolor}{RGB}{0,0,128}    
\definecolor{delcolor}{RGB}{128,0,0}    

\setaddedmarkup{\color{addcolor}#1}
\setdeletedmarkup{\color{delcolor}\sout{#1}}
\title{ProtAlign: Contrastive learning paradigm for Sequence and structure alignment}
\name{ Aditya Ranganath$^{*}$, Hasin Us Sami$^{*}$, Kowshik Thopalli, Bhavya Kailkhura, Wesam Sakla \thanks{$^{*}$ Equal contribution}\thanks{This work was performed under the auspices of the U.S. Department of Energy by Lawrence Livermore National Laboratory under contract DE-AC52-07NA27344.}}
\address{Center for Applied Scientific Computing, Lawrence Livermore National Laboratory}

\begin{document}
%
\maketitle
\begin{abstract}
Protein language models often take into consideration the alignment between a protein sequence and its textual description. However, they do not take structural information into consideration. Traditional methods treat sequence and structure separately, limiting the ability to exploit the alignment between the structure and protein sequence embeddings. In this paper, we introduce a sequence–structure contrastive alignment framework, which learns a shared embedding space where proteins are represented consistently across modalities. By training on large-scale pairs of sequences and experimentally resolved or predicted structures, the model maximizes agreement between matched sequence–structure pairs while pushing apart unrelated pairs. This alignment enables cross-modal retrieval (e.g., finding structural neighbors given a sequence), improves downstream prediction tasks such as function annotation and stability estimation, and provides interpretable links between sequence variation and structural organization. Our results demonstrate that contrastive learning can serve as a powerful bridge between protein sequences and structures, offering a unified representation for understanding and engineering proteins.
\end{abstract}
\section{Introduction}
Understanding the relationship between a protein’s sequence and its three-dimensional structure is a fundamental problem in computational biology~\cite{seqstructrelation}. In recent years, advances in machine learning and deep learning have driven striking improvements in protein structure prediction and functional reasoning from sequence information~\cite{af3}. While these advances largely focus on predicting structures from sequences, more recent trends extend the scope of protein representation learning to a modality alignment strategy. A prominent direction is to pair protein sequences with textual modalities, where natural language descriptions provide functional or contextual information that describes or complements sequence data~\cite{protst}. Consequently, these multi-modal \texttt{protein language models} have opened new avenues for linking sequence with higher-level biological knowledge. Approaches such as EvoLlama ~\cite{evollama} incorporate both sequence and structural information through pretrained encoders and subsequently integrate with an LLM like Llama~\cite{llama}.

\smallskip

Although multi-modal approaches have begun to incorporate both sequence and structure, they often rely on rudimentary operations such as concatenation or joint modeling, in lieu of explicit alignment between the two representations in a shared space. For example, existing methods either concatenate embeddings from different modalities~\cite{evollama} or jointly consume them in downstream models~\cite{concat-alignment}, but without a correlating constraint between sequence and structure representations. The lack of alignment limits cross-modal retrieval and interpretability; however, related research indicates that alignment approaches can be highly effective in protein modalities such as sequence–text learning~\cite{protst, protclip}. 

\smallskip

\textit{This motivates our fundamental objective - how can protein sequences and structures be consistently aligned in a shared embedding space? }To explore this, we introduce ProtAlign, a framework that builds on the success of contrastive alignment paradigm popularized by OpenAI’s CLIP~\cite{clip}. ProtAlign uses ESM2 for sequence embeddings and Protein-MPNN for structure embeddings, projecting them into a common space with a multi-head attention mechanism. By simultaneously maximizing correlation between paired samples and divergence between unpaired samples, ProtAlign produces interpretable, unified embeddings that enable cross-modal retrieval, improving downstream prediction tasks. 
\begin{figure*}[t]
    \centering
    \begin{tabular}{cc}
           \adjincludegraphics[width=0.44\textwidth, trim={6cm 7cm 10cm 5cm}, clip]{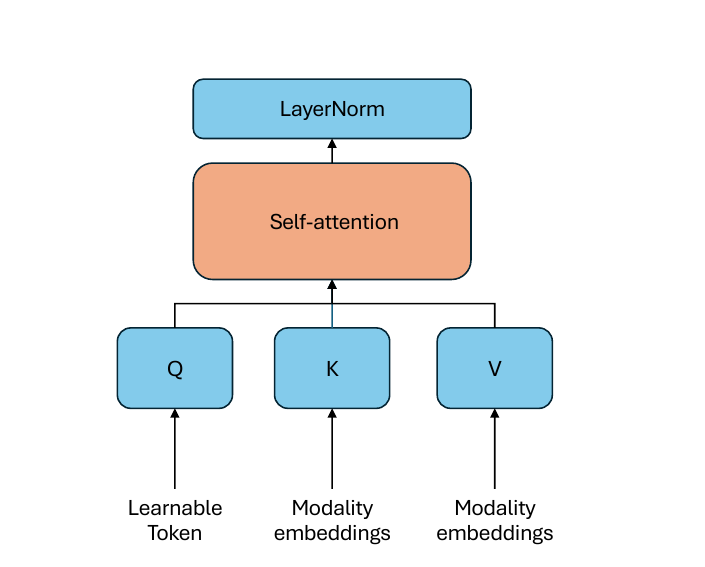} &\hspace{-10mm}\adjincludegraphics[width=0.44\textwidth, trim={11cm 6cm 3cm 4cm}, clip]{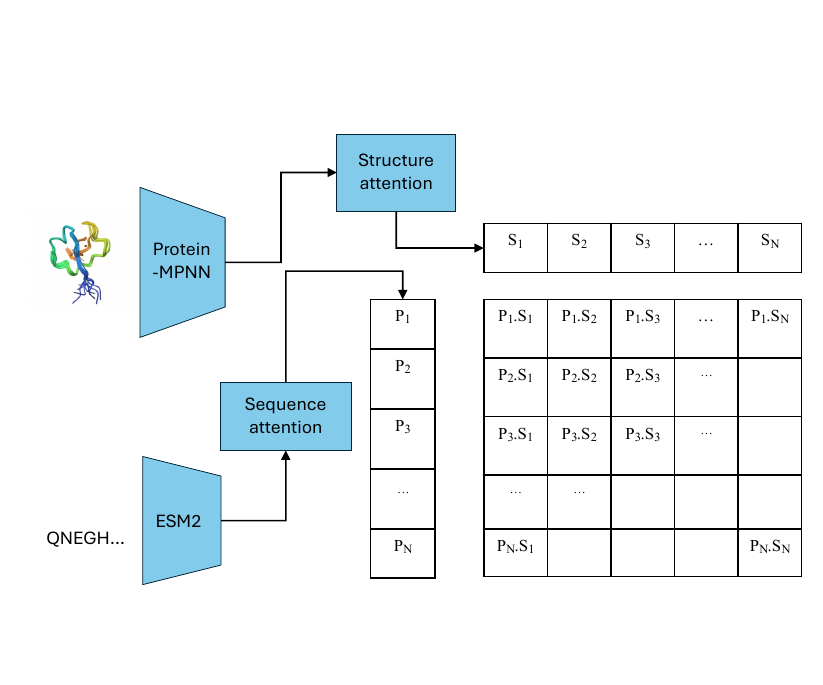} \\
           (a) Sequence/Structure model. & (b) Contrastive learning.
    \end{tabular}

    \caption{{\bf ProtAlign}.~ Fig.~(a) shows our proposed model consisting of multi-head self-attention (MSA) layer and LayerNorm layer, with the learnable token as queries, structure/sequence embeddings as keys and values within the MSA layer. Fig.~(b) shows the alignment protocol for training our model.}
    \label{fig:framework}
\end{figure*}
Through our experiments on the challenging PDBBind dataset, we observe that ProtAlign achieves a Recall@5 of 99.1\% (cross-modal retrieval) and yields interpretable embeddings that cluster structurally similar sequences. Beyond raw performance, we provide a holistic study of design choices that include choice of loss functions, temperature scaling, and projection strategies to better understand how alignment can be effectively achieved for protein data. Our codes will be released upon acceptance. 
\label{sec:intro}

\section{Proposed Approach}
In this section, we describe the working of the ProtAlign model in detail. For a given protein sequence and structure pair in our dataset, we feed the 3D structure to the Protein-MPNN \cite{proteinmpnn} encoder and retrieve a sequence of structure embeddings. Similarly, we feed the protein sequence to the ESM2 \cite{esm2} sequence encoder and retrieve a sequence of embeddings. We denote the sequence embeddings as $\mathbf{z}_P \in \mathbb{R}^{t_P \times D_P}$, where $D_P$ is the embedding dimension of ESM2 encoder and $t_P$ is the number of sequence tokens, and structure embeddings as $\mathbf{z}_S \in \mathbb{R}^{t_S \times D_S}$, where $D_S$ is the embedding dimension of the ProteinMPNN encoder output and $t_s$ is the number of structure tokens. For the rest of the paper, we consider $D \triangleq D_P$.

\begin{figure*}
\centering
\begin{tabular}{cc}
    \subfigure[Before-training.]{
    \includegraphics[width=0.35\linewidth, trim={1.3cm 1cm 0 0}, clip]{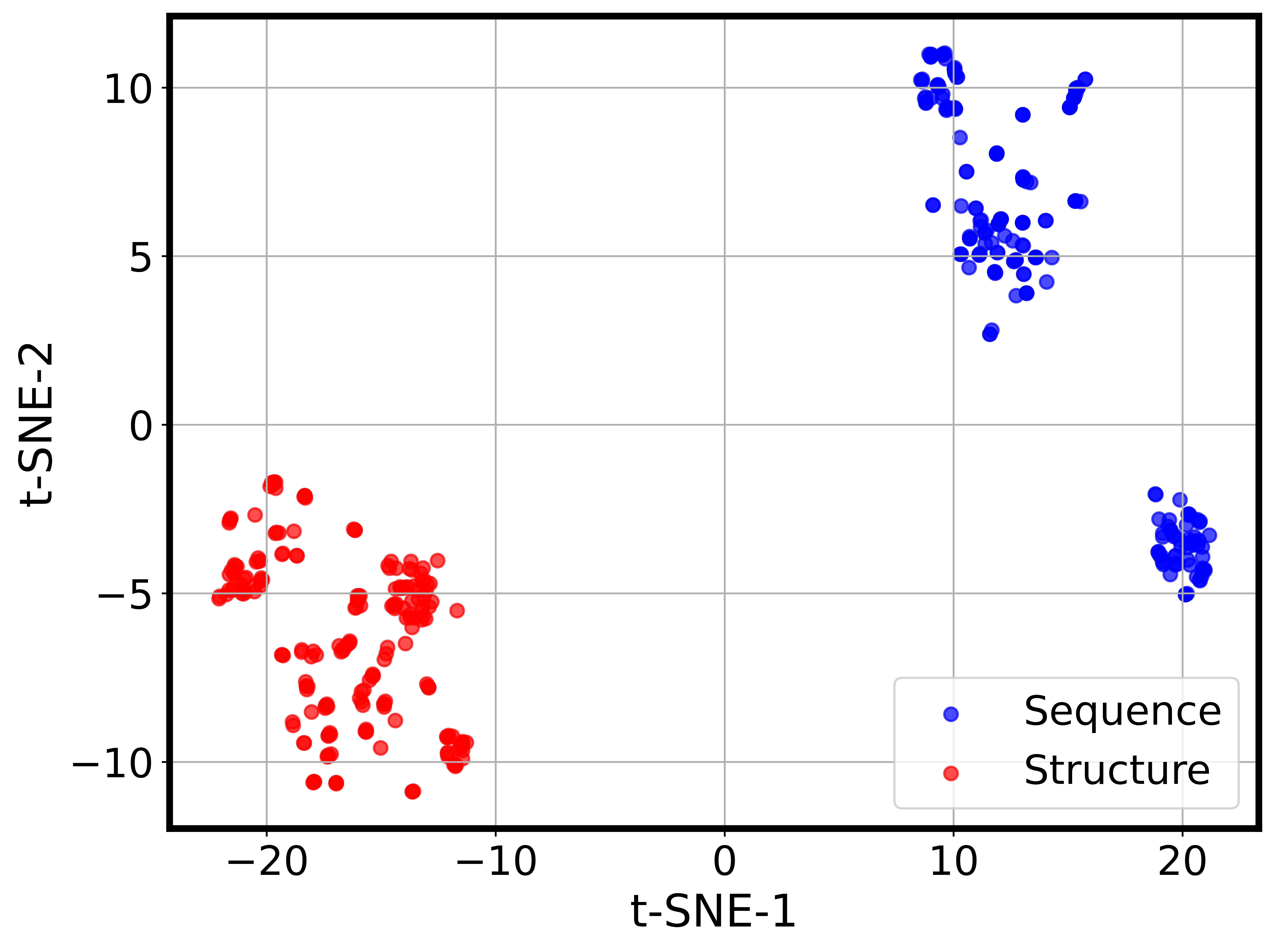}
    } 
    &
    \subfigure[Post-training.]{
        \includegraphics[width=0.36\linewidth, trim={1.3cm 1.5cm 0 0}, clip]{Figures/tSNE_t_0.07_fixed_marked.png}
    }
\end{tabular}
    \caption{The figure presents the t-SNE plot. Fig.\ (a) presents the 2D projection of the embeddings from both the protein and structure embeddings pre-training while Fig.\ (b) shows the t-SNE plot post-training. }
    \label{fig:tSNE_plot}
\end{figure*}

\noindent
{\bf Model architecture.} For the proposed model, we design a pair of multi-head self-attention (MSA) layers \cite{Vaswani2017} - one for each modality. We introduce two learnable tokens $\mathbf{z}_{P}^{Q}, \mathbf{z}_{S}^{Q} \in \mathbb{R}^D$, which enable us to a).\ project the sequence of tokens from the ESM2 embeddings and ProteinMPNN embeddings to a unified space and b).\ perform a weighted sum across the sequence space, extracting attention from the sequence of structure embeddings and sequence embeddings. We feed the learnable token $\mathbf{z}_P^{Q}$ as query $\mathbf{Q}_P$ and $\mathbf{z}_P$ as keys $\mathbf{K}_P$ and values $\mathbf{V}_P$.

The output from the MSA layer is fed to a LayerNorm (LN) \cite{Ba2016}, yielding
\begin{align}
    \mathbf{P} = \texttt{LN}\big(\texttt{softmax}\big(\frac{\mathbf{Q}_{P}^{\top} \mathbf{K}_{P}}{\sqrt{D}}\big) \mathbf{V}_{P}^{\top}\big)
\end{align}
as the final sequence embedding. In a similar fashion, we obtain our structure embedding $\mathbf{S}$. We use these sequence and structure embeddings to compute the CLIP \cite{clip} and SigLIP \cite{siglip} loss. For more details on the framework and the model architecture, please refer \cref{fig:framework}.






\noindent
{\bf CLIP \cite{clip}.} The CLIP loss, built upon the principle of softmax-based loss, can be written as
\begin{align}
\label{eq:clip}
\mathcal{L}_{\texttt{CLIP}} &\triangleq -\frac{1}{2N} \sum\limits_{i=1}^{N} \log \frac{\exp (\mathbf{P}_i \cdot \mathbf{S}_i/\tau)}{\sum_{j=1}^N \exp(\mathbf{P}_i\cdot \mathbf{S}_j/\tau)\}}\notag\\
&-\frac{1}{2N} \sum\limits_{i=1}^{N} \log \frac{\exp (\mathbf{P}_i \cdot \mathbf{S}_i/\tau)}{\sum_{j=1}^N \exp(\mathbf{P}_j\cdot \mathbf{S}_i/\tau)\}},
\end{align}
where temperature $\tau$ is used to control the sharpness of prediction probabilities and $N$ is the batch-size.

\noindent
{\bf SigLIP \cite{siglip}.} The SigLIP loss, which turns the alignment into a binary classification problem, can be written as
\begin{equation}
\label{eq:siglip}
\mathcal{L}_{\texttt{SigLIP}} \triangleq -\frac{1}{N} \sum\limits_{i=1}^{N} \sum\limits_{j=1}^{N} \log \frac{1}{1+ \exp\{y_{ij} (-\mathbf{P}_i\cdot \mathbf{S}_j/\tau + b)\}},
\end{equation}
where $y_{ij}=1$ if sequence $i$ and structure $j$ are paired and $-1$ otherwise. In addition to temperature $\tau$, SigLIP introduces a learnable bias term, $b$ to prevent heavy over-correction resulting from negative pairs in the loss function.

\label{sec:proposed}

\section{Experiments}

\begin{table*}[t]
    \centering
    \renewcommand{\arraystretch}{1.1}
    \caption{Protein sequences corresponding to samples within the highlighted region in the t-SNE plot (Fig.~\ref{fig:tSNE_plot}b). Differences are highlighted in \textcolor{red}{red}, illustrating how the proposed approach clusters highly similar protein sequences.}
    \begin{tabular}{|c|p{0.8\textwidth}|}
    \hline
    \rowcolor{gray!20}
    \textbf{PDB ID} & \textbf{Protein Sequence} \\
    \hline
    \texttt{3ao4} & {\tiny\ttfamily\raggedright
    SPGIWQLDCTHLEGKVILVAVHVASGYIEAEVIPAETGQETAYFLLKLAGRWPVKTVHTDNGSNFT
    \textcolor{red}{GATVRAACDWAGIKQE}
    IESMNKELKKIIGQVRDQAEHLKTAVQMAVFIHNHKRKG\textcolor{red}{GIGG}
    YSAGERIVDIIATD} \\
    \hline
    \texttt{3zso} & {\tiny\ttfamily\raggedright
    SPGIWQLDCTHLEGKVILVAVHVASGYIEAEVIPAETGQETAYFLLKLAGRWPVKTVHTDNGSNFT
    \textcolor{red}{STTVKAACWWAGIKQEDGIPYNPQSQGV}
    IESMNKELKKIIGQVRDQAEHLKTAVQMAVFIHNHKRKGYSAGERIVDIIATD
    \textcolor{red}{IQT}} \\
    \hline
    \texttt{3zsx} & {\tiny\ttfamily\raggedright
    SPGIWQLDCTHLEGKVILVAVHVASGYIEAEVIPAETGQETAYFLLKLAGRWPVKTVHTDNGSNFT
    \textcolor{red}{STTVKAACWWAGIKQEDGIPYNPQSQGV}
    IESMNKELKKIIGQVRDQAEHLKTAVQMAVFIHNHKRKGYSAGERIVDIIATD
    \textcolor{red}{IQ}} \\
    \hline
    \texttt{4cig} & {\tiny\ttfamily\raggedright
    SPGIWQLDCTHLEGKVILVAVHVASGYIEAEVIPAETGQETAYFLLKLAGRWPVKTVHTDNGSNFT
    \textcolor{red}{STTVKAACWWAGIKQEDGIPYNPQSQGV}
    IESMNKELKKIIGQVRDQAEHLKTAVQMAVFIHNHKR\textcolor{red}{G}
    YSAGERIVDIIATD\textcolor{red}{I}} \\
    \hline
    \end{tabular}
    \label{tab:sequence}
\end{table*}

\begin{table}[htpb]
    \centering
    \caption{Comparison of best cross-modal retrieval performance achieved by ProtAlign using SigLIP and CLIP loss functions. For CLIP, we set the temperature $\tau=0.07$ and for SigLIP, we set the bias $b=-10$.}
    \begin{tabular}{|c|c|c|c|}
    \hline
    \rowcolor{gray!20}
         \textbf{Method}  & \textbf{Recall@1 (\%)} & \textbf{Recall@5 (\%)} \\
    \hline
         SigLIP  &40.0 & 97.6 \\
         \hline
         CLIP  & \cellcolor{green!20}\textbf{42.7} & \cellcolor{green!20}\textbf{99.1} \\
    \hline
    \end{tabular}
    
    \label{tab:clip-siglip}
\end{table}

         
In this section, we describe the datasets, evaluation criteria, hyperparameter choices and empirical results. 

\noindent \textbf{Dataset.} To train the proposed ProtAlign paradigm, we use the PDBBind dataset \cite{pdbbind2020} as our dataset. The pdbbind dataset is broadly divided into 3 sub-categories - general, refined and CASF-2016. The general set, often reserved for training, contains $14,127$ protein-ligand complexes. The refined set contains $5,316$ complexes, while the core-set, or CASF-2016 subset contains $285$ complexes. Our main motivation for using this dataset owes to their experimentally solved 3D structures. Since we are only concerned with the sequences, we only use the protein sequences and discard the ligand SMILES. After removing the duplicates from all the sequences, the train set contains $10,071$, validation contains $3,387$ and test set contains $215$ sequences. 

\noindent\textbf{Evaluation.}  
We assess the effectiveness of the proposed approach on the sequence-to-structure retrieval task. Performance is measured using Recall, defined as the fraction of sequences for which the correct structure appears among the retrieved candidates. Specifically, we report Recall@1 and Recall@5, where Recall@$K$ denotes the fraction of sequences whose correct structure is found within the $K$ nearest structure embeddings ranked by cosine similarity.

\noindent
{\bf Hyperparameters.} We set the batch size, $N=1024$. The embedding dimension used for alignment is set to $D=128$. For training, we use Adam optimizer with a learning rate $\eta=0.001$. Within the MSA layer, we use $L=4$ heads.
\subsection{Results}
\begin{figure}[htp]
    \centering
    
    \adjincludegraphics[width=0.65\columnwidth]{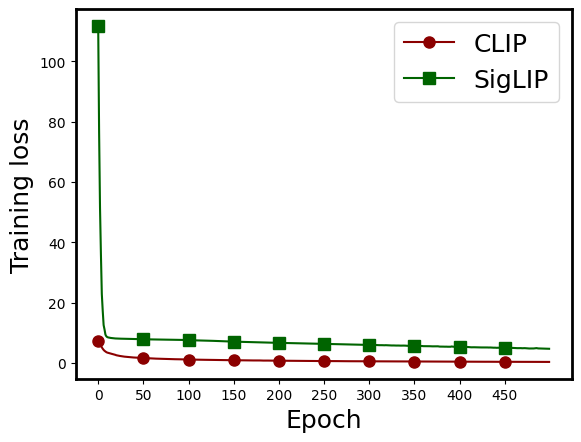}
    \caption{Training loss over epochs for CLIP vs. SigLIP.}
    \label{fig:training_loss}
    \vspace{-3pt}
\end{figure}

\smallskip
\noindent\textbf{How does the choice of loss function affect alignment?}
We first compare ProtAlign trained with CLIP-style contrastive loss~\cite{clip} against SigLIP~\cite{siglip} which are given by Eq~\ref{eq:clip} and Eq~\ref{eq:siglip} respectively. 
As shown in Table~\ref{tab:clip-siglip}, CLIP achieves a Recall@1 of 42.7\% and Recall@5 of 99.1\%, whereas SigLIP yields lower performance (40.0\% and 96.7\%, respectively). Moreover, training with CLIP converges faster and stabler than SigLIP (Figure~\ref{fig:training_loss}).  
We hypothesize that CLIP performs better in this setting due its softmax-based objective leveraging all negatives in the batch and directly optimizing relative similarity rankings. This is particularly beneficial when many proteins share partial similarities, since the model learns to distinguish fine-grained structural relationships. In contrast, SigLIP frames alignment as a binary classification task, which may penalize near-misses and converge more slowly.
From a biological perspective, protein families often consist of sequences that are highly similar and fold into nearly identical structures. In such cases, retrieving a  close ``structural neighbor'' can still be useful, even if it is not the ground-truth structure. CLIP’s ranking-based formulation naturally accommodates this by rewarding high similarity among clusters of related proteins, whereas SigLIP’s rigid binary separation may not fully capture the graded nature of sequence–structure relationships. Note, we also varied the bias term $b$ in the SigLIP loss function and observed that the best performance was achieved at $b = -10$. Given the improved performance of CLIP against SigLIP for this task, we adopt CLIP as the default loss function for ProtAlign in subsequent experiments.

\noindent\textbf{How does the temperature parameter in CLIP loss influence performance?}
We investigate the impact of the temperature parameter $\tau$ in the CLIP loss, which controls the sharpness of the similarity distribution. Table~\ref{tab:ablation-clip} summarizes Recall@1 and Recall@5 across different values of $\tau$. We find that $\tau = 0.07$ yields the best trade-off, achieving Recall@5 of 99.1\%. In contrast, very small values of $\tau$ (e.g., 0.001) lead to unstable training and degraded Recall@5.

\vspace{5mm}

\begin{table}[h]
    \centering
        \caption{We present the results for different temperature values in our CLIP loss. We notice that the temperature parameter $\tau =0.07$ yields the best results.}
    \begin{tabular}{|c|c|c|}
    \hline
    \rowcolor{gray!20}
         \textbf{Temperature} ($\tau$)&  \textbf{Recall@1} (\%) & \textbf{Recall@5} (\%) \\
         \hline
         \cellcolor{green!20}\textbf{0.07} & \cellcolor{green!20}\textbf{42.7} & \cellcolor{green!20}\textbf{99.1} \\
         0.035 & 42.3 & 98.6 \\
         0.02 & 42.7 & 98.6 \\
         0.001 & 40.93 & 95.8\\
         \hline
         
    \end{tabular}

    \label{tab:ablation-clip}
\end{table}

\begin{figure}[t]
\centering
\includegraphics[width=0.65\linewidth]{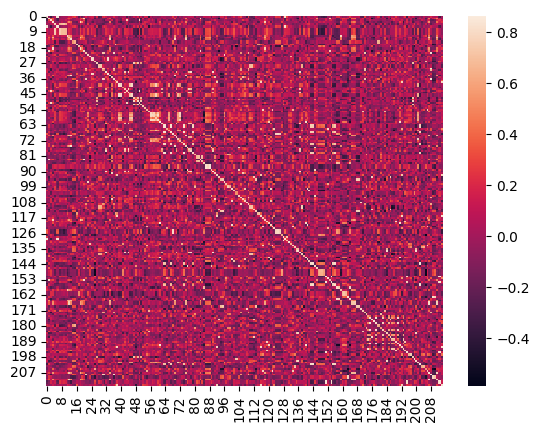}
\caption{Heatmap visualization of cosine similarity between all possible sequence-structure pairs ($\tau=0.07$). $y$-axis represents the sequences and $x$-axis represents the structures.}\label{fig:heatmap}
\end{figure}

\noindent\textbf{What do the embeddings in aligned space reveal qualitatively?}
To better understand the learned representation space, we visualize the sequence and structure embeddings using t-SNE~\cite{Maaten2008}. As shown in ~\cref{fig:tSNE_plot}, the embeddings before training with the contrastive loss are scattered without clear organization, whereas after training, ProtAlign produces well-defined clusters that bring sequences and their corresponding structures together. Interestingly, the model not only aligns individual sequence-structure pairs, but also groups families of related proteins into coherent neighborhoods. We highlight one such cluster in ~\cref{tab:sequence}, where the sequences of that cluster exhibit high similarity to each other. This behavior suggests that even when the retrieved structure is not the exact ground-truth pair it often corresponds to a protein with nearly identical sequence characteristics. From a domain-specific perspective, such learned latent space is meaningful as approximate neighbors can still yield meaningful functional or structural insights. Consequently, Recall@5 emerges as an appropriate metric, as it reflects the utility of retrieving close structural matches. We visualize cosine similarity scores between all sequence–structure pairs in the test set (~\cref{fig:heatmap}). The heatmap shows a strong diagonal dominance after alignment, reiterating that ProtAlign successfully maps matching pairs closer in the shared space.

\label{sec:results}

\section{Conclusion}
We introduced a novel alignment paradigm that unifies protein sequences and their corresponding structures within a shared embedding space. By leveraging complementary representations from sequence and structure modalities, our approach enables more effective multimodal learning. Through extensive and rigorous empirical experiments, we demonstrated the efficacy and robustness of this paradigm across a wide range of hyperparameter settings, underscoring its stability and generalizability. Beyond strong empirical performance, our framework provides a foundation for improving cross-modal retrieval, interpretability, and representation quality in protein modeling tasks. We believe this work opens new opportunities for integrating diverse biological modalities, paving the way for advances in downstream applications such as structure-based design and therapeutic discovery.
\label{sec:conclusion}

\bibliographystyle{IEEEbib}
\bibliography{strings,refs}

\end{document}